\documentclass[11pt]{article}

\usepackage[preprint]{acl}

\usepackage{times}
\usepackage{latexsym}
\usepackage[T1]{fontenc}
\usepackage[utf8]{inputenc}
\usepackage{microtype}
\usepackage{inconsolata}
\usepackage{graphicx}
\usepackage{amsmath}
\usepackage{amssymb}
\usepackage{booktabs}
\usepackage{multirow}
\usepackage{makecell}
\usepackage{array}
\usepackage{tabularx}

\title{When Agents Disagree: Bayesian Backward Reasoning as a Label-Free Anchor for Multi-Agent Collective Decision-Making}

\author{
  Ken Chen\textsuperscript{1}
  \quad Wei Wang\textsuperscript{1}
  \quad Sachith Seneviratne\textsuperscript{1}
  \quad Hansani Weeratunge\textsuperscript{2}
  \quad Saman Halgamuge\textsuperscript{1} \\
  \textsuperscript{1}Department of Mechanical Engineering, The University of Melbourne, Melbourne, Australia \\
  \textsuperscript{2}Department of Mechanical Engineering, Sri Lanka Institute of Information Technology, Sri Lanka \\
  \texttt{kenchen1@student.unimelb.edu.au}
}

\begin{document}
\maketitle

\begin{abstract}
When multiple LLM agents yield conflicting answers, the decision-making process dictates whether agent diversity improves performance or merely compounds shared errors. Existing collective decision-making methods, including voting, electoral rules, and LLM judges, rely on forward reasoning: they map evidence to labels in one direction. Although these methods can combine diverse forward traces, they still aggregate estimates that share this evidence-to-label factorization and can inherit correlated errors within the forward pool.
We therefore construct a reverse posterior for each instance through Bayesian backward reasoning from an explicit likelihood. The forward and reverse posteriors provide differently factorized approximations of the underlying posterior. Because estimates from different factorizations may tend to share the same error less often, we use Jensen-Shannon divergence to rank agents by cross-path consistency. This cross-path consistency signal underlies three strategies: hard selection (MinJS), soft reweighting (FwdJS), and log-linear fusion (LogLin).
Evaluated on DDXPlus across five LLM backbones, our proposed strategies show consistent improvements: MinJS outperforms random selection across all backbones, FwdJS generally improves over the strongest baseline, and LogLin achieves the best performance among the evaluated methods, with its largest gains on the subset where the agents disagree. Despite its weaker standalone accuracy, the reverse posterior serves as a more useful anchor than forward-only alternatives, providing complementary information for collective decision-making. When labeled data are available, a lightweight two-stage calibration can further refine the reverse anchor and improve aggregation performance.

\end{abstract}

\section{Introduction}
\label{sec:intro}

Large Language Model (LLM)-based multi-agent systems coordinate multiple agents to collectively solve complex tasks. One prominent paradigm extends single-model capabilities by prompting heterogeneous agents to address the same query, shifting the focus toward collaborative decision-making, where diverse outputs must be aggregated into a final consensus. This paradigm is rooted in human collective intelligence~\citep{woolley-etal-2010-evidence,zhou-etal-2026-identifying} and cognitive diversity, positing that a team of diversely skilled solvers can outperform a uniform group of top individuals~\citep{hong-page-2004-groups}. However, diversity alone does not guarantee superior group performance. When agents disagree, the aggregation mechanism ultimately determines whether complementary insights are synthesized or shared errors are compounded.

Existing collective decision-making pipelines typically rely on voting-based mechanisms~\citep{zhao-etal-2024-electoral} or a designated LLM judge~\citep{zhao-etal-2026-agentcdm}. However, these approaches present distinct limitations. Voting and electoral rules-based methods rely on simple consensus and lack a reliable, objective anchor to resolve conflicts when agents disagree. 
Conversely, dictatorial approaches such as LLM-as-a-judge attempt to supply an external decision, but the evaluation remains grounded in the same forward-generated traces.
Furthermore, they have also been observed to exhibit systematic biases, such as position bias, verbosity bias, or self-enhancement bias~\citep{zheng-etal-2023-judging}. While multi-round debate can refine opinions~\citep{du-etal-2024-improving,liang-etal-2024-encouraging}, it incurs high computational costs and remains vulnerable to conformity, identity-driven sycophancy, and persona instability~\citep{baltaji-etal-2024-conformity,choi-etal-2026-identity,li2026single}.

Focusing on single-round aggregation, we argue that the critical missing ingredient is a reliable external anchor, a reference distribution that guides the system in selecting and fusing conflicting answers. The root cause of aggregation failure is directional rather than merely algorithmic. Plurality voting, electoral rules, and LLM judges all operate exclusively within \emph{forward reasoning}: they reason directly from evidence to conclusions along the exact same conditioning path as the agents they evaluate. Consequently, errors within this pool are highly collinear. If the majority hallucinates, a vote or a judge is likely to echo the same mistake, trapping the system in a collective error. To break this cycle, the system requires a reference that resides in the same class-posterior space but is differently factorized, constructed via a path distinct from standard forward elicitation.

Bayes' theorem provides two complementary factorizations of the same posterior. While forward reasoning executes one as a discriminative evidence-to-conclusion mapping, we explicitly construct the other through \emph{Bayesian backward reasoning}. By defining an explicit likelihood $P(e\mid d)$ over evidence $e$ and labels $d\in\mathcal{Y}$, together with a prior, we invert the inference once per instance, a process of Bayes-style backward reasoning that yields a shared reverse posterior $R(d\mid x)$, where $x$ denotes the observed input. Consequently, each agent's forward posterior $F_i$ and $R$ serve as differently factorized, biased approximations of the same class posterior $P(d\mid x)$. Our working hypothesis is that their respective errors are far less collinear than those strictly within the forward pool. Because biased estimates derived from different factorizations are less likely to converge on the same incorrect conclusion than estimates from the same factorization, cross-path agreement provides a useful consistency signal for identifying more reliable predictions. Therefore, we utilize the Jensen-Shannon divergence, $\mathrm{JS}(F_i, R)$, as a ranking signal for cross-path consistency rather than a definitive certificate of correctness. This symmetric metric treats neither channel as absolute ground truth, and its bounded nature ensures comparability across instances. Crucially, measuring distance to the pool's internal consensus, such as the majority vote or the average prediction, cannot fulfill this role: it merely rewards conformity, actively penalizing the rare agent that happens to be correct when the majority is wrong.

Figure~\ref{fig:overview} summarizes our reverse-anchored framework: a shared reverse posterior serves as a reference for selecting, reweighting, and fusing the forward agents.
Concretely, a single backward inference generates the shared reverse posterior $R$ via explicit Bayesian backward reasoning over the candidate shortlist proposed by the agents’ forward reasoning, while each $F_i$ represents an agent's forward posterior. Unlike existing backward checks that merely score individual reasoning chains in math problems~\citep{weng-etal-2023-large,jiang-etal-2024-forward}, our $R$ serves as a complete class distribution used to evaluate and fuse multiple forward agents. We leverage the Jensen--Shannon divergence, $\mathrm{JS}(F_i,R)$, through three progressive aggregation strategies:
\begin{itemize}
    \item \textbf{MinJS (Hard Selection):} Selects the single forward agent closest to $R$. While the ranking is informative, hard selection can be brittle because cross-path agreement measures compatibility with $R$, not an absolute guarantee of correctness.
    \item \textbf{FwdJS (Soft Reweighting):} Softly reweights the forward pool based on their alignment with $R$. Crucially, $R$ solely determines the aggregation weights, meaning the final prediction remains a pure mixture of the forward posteriors:  $R$ shapes the weights but contributes no probability mass of its own. 
    \item \textbf{LogLin (Log-Linear Fusion):} Actively blends $R$ into the final prediction as a multiplicative factor with a small fixed weight ($w_R{=}0.2$). This combination captures complementary cross-path information that the first two methods miss, as they rely solely on the forward outputs.
\end{itemize}

\paragraph{Our main contributions are summarized as follows:}
\begin{enumerate}
    \item \textbf{Cross-path reverse anchor.} We introduce a reverse posterior obtained by Bayesian backward reasoning that acts as a reliable, label-free reference for multi-agent decision-making. We demonstrate that $\mathrm{JS}(F_i,R)$ effectively ranks and routes conflicting agents based on cross-path consistency.
    \item \textbf{Label-free aggregation suite.} We propose three training-free strategies (MinJS, FwdJS, and LogLin). Extensive experiments demonstrate the progressive effectiveness of our proposed suite: MinJS consistently outperforms the selection baseline, FwdJS improves over the strongest baseline in most settings, and LogLin achieves strictly superior performance against all baselines across all scenarios.
    \item \textbf{Anchor indispensability.} We establish that despite often being the weakest standalone predictor, the reverse posterior $R$ acts as an irreplaceable anchor. Replacing it with the pool-mean forward posterior ($MeanF$) or a pool-external general single-agent forward posterior ($GenF$) strictly degrades both FwdJS and LogLin, confirming the unique value of cross-path diversity.
    \item \textbf{Lightweight labeled calibration.} When a limited labeled split is available, we apply a two-stage calibration that repairs the reverse anchor's quality and further elevates the aggregation strategies.
\end{enumerate}

\section{Related Work}
\label{sec:related}

\paragraph{Multi-agent collective decision-making.}
LLM multi-agent systems commonly coordinate through debate, role specialization, or layered synthesis~\citep{du-etal-2024-improving,liang-etal-2024-encouraging,wang-etal-2025-mixture}. Once multiple agents produce answers to the same instance, the system must aggregate their individual decisions into a single collective outcome. Existing approaches largely do so either by aggregating agents' ballots through majority voting or more elaborate electoral rules~\citep{zhao-etal-2024-electoral,ai-etal-2025-beyond}, or by delegating the decision to a designated judge~\citep{zheng-etal-2023-judging,liu-etal-2023-g,chan-etal-2024-chateval,zhao-etal-2026-agentcdm}. Confidence-weighted consensus further refines the former through discussion~\citep{chen-etal-2024-reconcile}, while recent analyses suggest that simple majority already captures much of the benefit attributed to multi-round debate~\citep{choi-etal-2025-debate}. Despite these differences, both voting and judging derive their decision signal from the agents' forward outputs. In contrast, we derive an instance-level reference from a generative reverse posterior, providing a training-free alternative to another vote or judge.

\paragraph{Forward--backward reasoning.}
A separate line of work uses backward reasoning to evaluate a proposed answer or label rather than to aggregate decisions from multiple agents. Self-Verification~\citep{weng-etal-2023-large} introduces backward checks at inference time to assess candidate solutions by testing whether the candidate is consistent with the underlying conditions. FOBAR~\citep{jiang-etal-2024-forward} further combines Self-Consistency-style forward answer votes~\citep{wang-etal-2023-self} and backward probabilities through a geometric mean. RevThink~\citep{chen-etal-2025-reverse} instead trains a model to internalize forward and backward reasoning, while inference still produces a single forward answer. These methods thus use reverse reasoning primarily as a candidate-level verification signal. Rather than using the reverse signal only to verify individual candidate-level predictions, we derive a full reverse posterior and use its cross-path consistency with the forward posteriors to guide selection, reweighting, and fusion across multiple agents.

\paragraph{Ensembling and unlabeled routing.}
Model ensembling and routing also combine information from multiple models, but they formulate the problem as prediction fusion or model selection rather than collective decision-making among explicit agents. LLM-Blender~\citep{jiang-etal-2023-llm} merges candidate answer texts. DeePEn~\citep{huang-etal-2024-ensemble} and PackLLM~\citep{mavromatis-etal-2024-pack} combine next-token predictions across models. SMOOTHIE~\citep{guha-etal-2024-smoothie} performs unlabeled routing by scoring each model’s sample-wise quality from the unlabeled outputs and sending the input to the highest-scoring model. These methods operate on forward predictions and use agreement or predictive compatibility as their quality signal. In contrast, we retain the multi-agent collective-decision setting and introduce a reverse posterior as an external reference without relying on another forward reasoning agent or a learned judge.

\section{Method}
\label{sec:method}

\begin{figure*}[t]
  \centering
  \includegraphics[width=\textwidth]{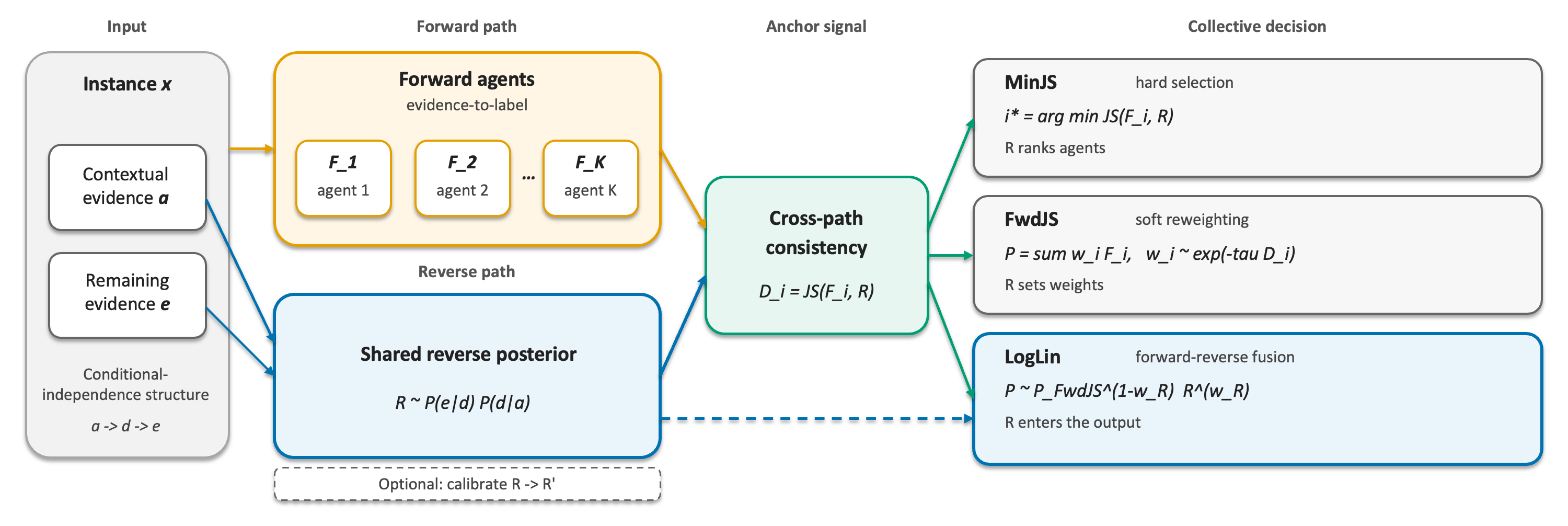}
  \caption{Reverse-anchored aggregation, left to right. Input $x$ is decomposed into contextual evidence $a$ and remaining evidence $e$ under the structure $a{\rightarrow}d{\rightarrow}e$. The forward path (yellow) produces agent posteriors $F_i$, while the reverse path (blue) produces a shared reverse posterior $R\propto P(e\mid d)\,P(d\mid a)$. The shared anchor signal (green), $D_i=\mathrm{JS}(F_i,R)$, guides all three heads: MinJS selects one $F_i$, FwdJS reweights $\{F_i\}$, and LogLin additionally incorporates $R$ into the output. The dashed box denotes optional labeled calibration $R{\to}R'$. The dashed blue arrow indicates the direct contribution of $R$ to LogLin.}
  \label{fig:overview}
\end{figure*}

\subsection{Collective decision over agent posteriors}
\label{sec:setup}

Consider an instance $x$ and a finite label set $\mathcal{Y}$, shown at the left of Figure~\ref{fig:overview}.
$K$ heterogeneous agents $\mathcal{A}={A_1,\ldots,A_K}$ each answer the same $x$ and return a forward posterior $F_i(d\mid x)$ over $d\in\mathcal{Y}$.
Our goal is to make a \emph{collective decision} from the agent posteriors ${F_i}_{i=1}^{K}$.
We consider two forms of collective decision: \textbf{selection}, which adopts the posterior of one agent, and \textbf{fusion}, which fuses the pool into a single posterior $P(d\mid x)$ and predicts
$\arg\max_d P(d\mid x)$.
Both are implemented within the same label-free and training-free inference framework.

\subsection{Reverse posterior as a consistency anchor}
\label{sec:anchor}
The class label $d\in\mathcal{Y}$ is latent; what differs across inference paths is \emph{how} the evidence is conditioned on $d$. Forward agents elicit $F_i(d\mid x)$ through a discriminative route from the instance $x$ to the label space.
We instead construct a complementary reverse view by specifying how the observed evidence factorizes conditioned on $d$ and then applying Bayes' rule.

We decompose the input $x$ into two evidence components, $a$ and $e$, and posit the conditional-independence structure $a\rightarrow d\rightarrow e$.
Here, $a$ represents contextual evidence, while $e$ provides additional evidence about the latent label $d$.
This implies $P(e\mid a,d)=P(e\mid d)$, and Bayes' rule gives
\begin{equation}
P(d\mid a,e)
\propto
P(e\mid d)\,P(d\mid a).
\end{equation}

We thus obtain a \emph{reverse likelihood} $P(e\mid d)$ and a label prior $P(d\mid a)$, and define the resulting \emph{reverse posterior}
\begin{equation}
R(d\mid x)
\propto
P(e\mid d)\,P(d\mid a),
\label{eq:reverse}
\end{equation}
normalized over $\mathcal{Y}$.
Forward and reverse inference therefore provide two different conditionalization perspectives on the same latent label: direct evidence-to-label prediction versus generative inversion.

Neither path is expected to recover the exact posterior $P(d\mid x)$. LLM elicitation and model bias make both $F_i$ and $R$ \emph{biased approximations} rather than ground-truth posteriors. In particular, no exact Bayes identity links an elicited forward posterior $F_i$ to our constructed reverse posterior $R$.
The motivation for introducing $R$ is weaker and does not require either estimate to be exact: forward agents can exhibit correlated errors due to shared model priors or reasoning patterns, while the reverse factorization provides a structurally different source of evidence. We therefore do not treat $R$ as a ground-truth posterior, but as a \emph{structurally distinct consistency signal} whose errors need not be fully aligned with those of the forward predictors.
This motivates using $R$ to rank or weight the forward posteriors rather than introducing another forward vote or judge.

We instantiate the reverse anchor with one shared reverse inversion per instance, which uses ordinal likelihood and context-activation maps together with two LLM elicitation steps over the agents' top-$k$ labels, followed by exact replay on the Bayesian network.
The construction is an implementation of the reverse anchor rather than a requirement of the aggregation framework itself.
For the label-free experiments, the resulting reverse distribution is used without any training data.

Because the forward and reverse estimates arise from different factorizations, their errors can be partially complementary.
This motivates a \emph{product-like composition} that emphasizes labels supported by both channels, without requiring the stronger assumption that either distribution is exact or that the two estimates are statistically independent.

\subsection{Reverse-referenced selection and fusion}
\label{sec:aggregation}
We measure the consistency between each forward posterior and the reverse anchor (cross-path consistency) using the Jensen-Shannon divergence:
\begin{equation}
\begin{aligned}
D_i &= \mathrm{JS}(F_i,R), \\
    &= \tfrac12\mathrm{KL}(F_i\|M_i) + \tfrac12\mathrm{KL}(R\|M_i), \\
M_i &= \tfrac12(F_i+R).
\end{aligned}
\label{eq:js}
\end{equation}
The reverse posterior $R$ serves as a shared external reference, shown in green in Figure~\ref{fig:overview}, and the same consistency score is reused by the three heads on the right: hard selection, soft reweighting, and forward--reverse fusion.

\paragraph{MinJS (hard selection).}
MinJS selects the agent whose posterior is most consistent with the reverse anchor:
\begin{equation}
\begin{gathered}
i^{\star} = \arg\min_i D_i, \\
\hat{d} = \arg\max_d F_{i^{\star}}(d).
\end{gathered}
\label{eq:minjs}
\end{equation}
The output $\hat{d}$ is obtained from the selected agent's forward posterior $F_{i^{\star}}$.
Thus, $R$ is used only to rank the agents and does not contribute probability mass to the final prediction.

\paragraph{FwdJS (soft reweighting).}
MinJS uses reverse consistency for hard selection. FwdJS converts the same signal into soft weights over the forward pool:
\begin{equation}
\begin{gathered}
w_i = \frac{\exp(-\tau D_i)}{\sum_j \exp(-\tau D_j)}, \\
P_{\mathrm{FwdJS}}(d) = \sum_i w_i F_i(d).
\end{gathered}
\label{eq:fwdjs}
\end{equation}
We fix the temperature to $\tau=5.0$ throughout all experiments.
The collective label is $\arg\max_d P_{\mathrm{FwdJS}}(d)$.
Agents whose posteriors are more consistent with $R$ receive larger weights, while the resulting posterior remains a convex combination of the forward posteriors.
In this operator, $R$ shapes the routing weights but does not itself contribute probability mass to the output.

\paragraph{LogLin (forward--reverse fusion).}
When the forward and reverse estimates contain partially complementary errors, a product-like composition can emphasize labels supported by both channels.
LogLin therefore builds on the FwdJS aggregate and directly incorporates the reverse posterior:
\begin{equation}
P_{\mathrm{LogLin}}(d) \propto P_{\mathrm{FwdJS}}(d)^{\,1-w_R}\, R(d)^{\,w_R},
\label{eq:loglin}
\end{equation}
where we use $w_R=0.2$ by default unless stated otherwise (Appendix~\ref{sec:appendix-wr-sweep}).
FwdJS determines how much each forward agent contributes, whereas LogLin additionally allows the reverse anchor to reshape the final posterior (dashed arrow in Figure~\ref{fig:overview}).
This fixed-weight product-like form is related at the operator level to the geometric-mean composition used in FOBAR~\citep{jiang-etal-2024-forward}, but serves a different purpose: FOBAR combines forward and backward evidence for candidate verification, whereas we use the reverse posterior as a shared anchor for multi-agent aggregation.

\paragraph{Summary.}
The reverse posterior $R$ is not treated as another competing prediction.
Instead, it provides a shared consistency reference that supports three levels of collective decision: hard selection (MinJS), soft reweighting (FwdJS), and direct forward--reverse fusion (LogLin).
Importantly, cross-path consistency is a measure of compatibility rather than a certificate of correctness. Consequently, $R$ need not be the top-1 predictor itself to be useful as an aggregation anchor.

\subsection{Labeled calibration of the reverse anchor}
\label{sec:calib-method}

The ordinal likelihood and context-activation maps used to construct $R$ may be misspecified, making the resulting reverse posterior a biased approximation of the underlying posterior.
When a labeled training split is available, we optionally calibrate the reverse anchor $R$ while keeping MinJS, FwdJS, and LogLin fixed (dashed box in Figure~\ref{fig:overview}).
The low-capacity parameterization is designed to permit calibration with limited labeled data, without retraining the underlying LLM. Calibration only modifies the reverse anchor used by the aggregation heads. 

Stage~1 calibrates the two ordinal maps used to construct the reverse posterior: the ordinal likelihood map used to parameterize $P(e\mid d)$ and the contextual-activation map fused to parameterize $P(d\mid a)$.
Each map converts an ordinal rank $k\in\{0,\ldots,6\}$ into a monotone continuous value through a logit-linear curve:
\begin{equation}
\label{eq:l5curve}
v_k = \ell + (h-\ell)\,\sigma\!\bigl(a + b\cdot k/6\bigr), \quad b>0,
\end{equation}
where $\sigma$ is the logistic function and $(a,b)$ are learned separately for the two maps.
A temperature $T$ is additionally fitted to rescale the replayed reverse posterior.
After fitting, the two maps and $T$ are frozen, and the reverse posterior $R$ is recomputed by replaying the Bayesian network with the calibrated parameters.

Stage~2 then applies a scalar class-prior correction
\begin{equation}
\label{eq:l1g}
R'(d) \;\propto\; \frac{R(d)}{m(d)^{\gamma}},
\end{equation}
where $m(d)$ is the class marginal of $R$ and $\gamma$ is a single scalar parameter.
The resulting calibrated reverse anchor $R'$ therefore adds only this scalar correction and does not introduce a class-specific bias $b_d$.
Fitting details and a higher-capacity variant with a class-specific bias vector $b\in\mathbb{R}^{|\mathcal{Y}|}$, used as a capacity upper bound, are provided in Appendix~\ref{sec:appendix-l1gb}.

\subsection{Computational cost}
\label{sec:cost}

Inference requires $K$ forward agent inferences and one shared reverse inversion per instance, followed by lightweight divergence computation and posterior aggregation.
The reverse inversion is implemented with two LLM elicitation steps composed through the Bayesian network, and the same reverse posterior $R$ is reused by all aggregation operators.
No parameter updates are performed at test time; labeled calibration, when used, is fit on a training split only.
Thus, the method introduces no fine-tuning cost and requires only a single shared reverse construction in addition to the $K$ forward agent inferences.

\section{Experimental Setup}
\label{sec:exp-setup}

\subsection{Dataset}
\label{sec:dataset}

We evaluate on DDXPlus~\citep{fansi-tchameu-etal-2022-ddxplus}, a synthetic diagnostic benchmark whose cases pair structured clinical evidence with a closed set of 49 disease labels.
Each case records demographics and observed findings, together with contextual information that precedes the remaining evidence.
For this benchmark, the contextual component corresponds to $a$, the remaining observed evidence to $e$, and the latent class $d$ to the diagnosis.
Forward agents map the full case description to a disease posterior, while the reverse procedure constructs
$R(d\mid x)\propto P(e\mid d)\,P(d\mid a)$ once per case from the same evidence.
We use a fixed 2,000-case test slice.
Because every aggregator must use the same five forward posteriors and the same shared reverse anchor, we restrict evaluation to the per-backbone intersection of complete forward cases and cases with an available reverse posterior.
We apply the same availability and parsing criteria to all methods within each backbone, yielding a fixed evaluation pool independent of the aggregation method.
\emph{All} denotes this complete per-backbone pool.
\emph{Disagree} is the subset of \emph{All} in which the five forward top-1 predictions are not unanimous.
Headline tables use these two slices.

\subsection{Baselines}
\label{sec:baselines}

Collective-decision baselines operate entirely within the forward channel.
\emph{Random Agent (Rand.)} uniformly samples one of the five forward agents per case. We report the pooled accuracy over three fixed seeds. It serves as the selection control for MinJS.
Voting-based aggregators follow GEDI~\citep{zhao-etal-2024-electoral} and operate on the same five forward ballots or posteriors without reverse input.
Plurality counts top-1 votes; Range voting treats each posterior as a cardinal score vector and selects the disease with the highest total score; Borda Count, Bucklin, IRV, Minimax, and Ranked Pairs apply the corresponding ordinal scoring or runoff rules.
The dictatorial-based baselines are two LLM judges over the same five agents: Informed dictatorial judge (Informed Dicta.) and ACH-inspired structured judge~\citep{zhao-etal-2026-agentcdm}.

\subsection{Models and implementation details}
\label{sec:exp-protocol}

The forward pool comprises five heterogeneous prompting strategies applied to each backbone: Tree-of-Thought (ToT), few-shot MedPrompt, vanilla chain-of-thought, common-bias, and rare-bias.
Each agent produces a top-5 posterior over the official 49-label set.
We use unified sampling settings across all agents, with temperature $=0.6$ and top-$p=0.95$, and generate $n{=}5$ responses per query, one for each agent.
We evaluate the resulting pools on five backbones: Qwen3-30B-A3B~\citep{yang-etal-2025-qwen3} (hereafter Qwen3-30B), Mistral-Small-3.1-24B~\citep{mistral-ai-2025-small}, Llama-3.1-8B~\citep{grattafiori-etal-2024-llama}, GLM-4-32B~\citep{glm-etal-2024-chatglm}, and DeepSeek-V4-Flash~\citep{deepseek-ai-2026-v4}.
The primary metric is Gold Top-Pathology Accuracy at 1 (GTPA@1), where a prediction is counted as correct when the gold pathology is the $\arg\max$ of the final posterior.
The shared reverse posterior is the label-free inversion described in Section~\ref{sec:anchor} and is reused by all reverse-referenced operators for each case.
Our methods are MinJS selection, FwdJS reweighting with $\tau=5.0$, and LogLin forward--reverse fusion with fixed $w_R=0.2$ (Tables~\ref{tab:main-valid}--\ref{tab:main-disagree}; see Appendix~\ref{sec:appendix-wr-sweep} for the $w_R$ sweep).
Within each backbone, all methods are evaluated on the same fixed case set.

\section{Results}
\label{sec:results}
\label{sec:experiments}

\subsection{Main Results: Reverse-Anchored Collective Decision}
\label{sec:main-results}

A shared reverse posterior turns an unanchored forward pool into a cross-path collective decision.
Tables~\ref{tab:main-valid} and~\ref{tab:main-disagree} report GTPA@1 on \emph{All} and \emph{Disagree}.
A double rule separates single-agent selection from pool aggregation.
For selection methods, the best result is shown in bold; for voting, judging, and fusion methods, the best and runner-up results are shown in bold and underline, respectively.

\begin{table*}[t]
  \centering
  \scriptsize
  \setlength{\tabcolsep}{2.2pt}
  \begin{tabular}{l|cc||ccccccc|cc|cc}
    \toprule
    & \multicolumn{2}{c||}{Selection} & \multicolumn{7}{c|}{Voting-based} & \multicolumn{2}{c|}{Dictatorial-based} & \multicolumn{2}{c}{Our method} \\
    \cmidrule(lr){2-3} \cmidrule(lr){4-10} \cmidrule(lr){11-12} \cmidrule(lr){13-14}
    \textbf{Model} & \makecell{Rand.} & MinJS & Plurality & Range & \makecell{Borda\\Count} & Bucklin & IRV & Minimax & \makecell{Ranked\\Pairs} & \makecell{Informed\\Dicta.} & ACH & FwdJS & LogLin \\
    \midrule
    Qwen3-30B & 70.10 & \textbf{71.64} & 71.54 & 72.69 & 71.34 & 71.94 & 72.14 & 71.54 & 71.69 & 72.84 & 71.84 & \underline{73.85} & \textbf{74.25} \\
    Mistral-Small-3.1-24B & 69.18 & \textbf{71.37} & 71.37 & 73.20 & 72.54 & 72.19 & 72.29 & 72.03 & 72.29 & 72.39 & 72.13 & \underline{73.51} & \textbf{74.58} \\
    Llama-3.1-8B & 48.39 & \textbf{49.87} & 54.11 & 57.90 & 54.16 & 54.42 & 54.01 & 52.75 & 54.47 & 56.39 & 50.38 & \underline{58.15} & \textbf{58.46} \\
    GLM-4-32B & 62.16 & \textbf{64.26} & 65.32 & 65.12 & 64.31 & 65.42 & 64.82 & 64.66 & 65.57 & 65.93 & \underline{66.78} & 66.18 & \textbf{67.04} \\
    DeepSeek-V4-Flash & 75.35 & \textbf{77.60} & 76.84 & 77.80 & 77.60 & 77.29 & 77.19 & 76.99 & 77.14 & 76.58 & 75.87 & \underline{78.61} & \textbf{79.07} \\
    \bottomrule
  \end{tabular}
  \caption{GTPA@1 (\%) on \emph{All}.
  `Rand.' and `Dicta.' denote `random' and `dictatorial'. The double rule separates single-agent selection from pool aggregation. For selection, the best result is shown in bold; among pool-aggregation methods, the best and runner-up results are shown in bold and underlined, respectively.}
  \label{tab:main-valid}
\end{table*}

\begin{table*}[t]
  \centering
  \scriptsize
  \setlength{\tabcolsep}{2.2pt}
  \begin{tabular}{l|cc||ccccccc|cc|cc}
    \toprule
    & \multicolumn{2}{c||}{Selection} & \multicolumn{7}{c|}{Voting-based} & \multicolumn{2}{c|}{Dictatorial-based} & \multicolumn{2}{c}{Our method} \\
    \cmidrule(lr){2-3} \cmidrule(lr){4-10} \cmidrule(lr){11-12} \cmidrule(lr){13-14}
    \textbf{Model} & \makecell{Rand.} & MinJS & Plurality & Range & \makecell{Borda\\Count} & Bucklin & IRV & Minimax & \makecell{Ranked\\Pairs} & \makecell{Informed\\Dicta.} & ACH & FwdJS & LogLin \\
    \midrule
    Qwen3-30B & 40.35 & \textbf{45.26} & 44.96 & 48.12 & 44.06 & 45.86 & 46.47 & 44.96 & 45.41 & 48.57 & 46.17 & \underline{51.58} & \textbf{52.78} \\
    Mistral-Small-3.1-24B & 40.69 & \textbf{46.24} & 46.24 & 50.59 & 48.88 & 47.95 & 48.22 & 47.56 & 48.22 & 48.48 & 49.01 & \underline{51.39} & \textbf{53.90} \\
    Llama-3.1-8B & 35.64 & \textbf{37.76} & 44.14 & 49.70 & 44.14 & 44.52 & 43.92 & 42.04 & 44.59 & 47.60 & 40.62 & \underline{50.08} & \textbf{50.90} \\
    GLM-4-32B & 33.63 & \textbf{38.19} & 40.79 & 40.23 & 38.42 & 40.90 & 39.55 & 39.32 & 41.36 & 42.03 & \underline{44.07} & 42.60 & \textbf{44.41} \\
    DeepSeek-V4-Flash & 41.38 & \textbf{48.15} & 45.93 & 48.44 & 47.85 & 46.96 & 46.67 & 46.07 & 46.52 & 45.19 & 44.15 & \underline{50.81} & \textbf{52.30} \\
    \bottomrule
  \end{tabular}
  \caption{GTPA@1 (\%) on \emph{Disagree}.
  Column groups and formatting match Table~\ref{tab:main-valid}.}
  \label{tab:main-disagree}
\end{table*}

LogLin consistently outperforms the strongest electoral baseline.
On \emph{Disagree}, it exceeds the best forward-only electoral rule on all five backbones by $1.2$--$4.7$~pp and also outperforms both LLM judges.
On \emph{All}, which includes cases where the forward agents already agree, LogLin achieves the best result for every backbone. Gains on \emph{Disagree} are therefore diluted in the aggregate score.
FwdJS likewise outperforms the strongest electoral rule on \emph{Disagree} by $0.4$--$3.5$~pp.
The comparison on GLM illustrates that reverse-guided reweighting is not uniformly dominant: ACH reaches $44.07\%$, slightly above FwdJS at $42.60\%$, while LogLin remains best at $44.41\%$.

The three operators also exhibit the intended progression from hard selection to soft aggregation and direct reverse fusion.
MinJS already exploits reverse consistency, outperforming random agent on every backbone, but hard selection can remain below the strongest election (e.g., $71.64\%$ vs.\ $72.69\%$ for Range voting on Qwen \emph{All}).
FwdJS improves on this by distributing weight across agents according to their consistency with $R$, while LogLin further incorporates $R$ directly into the fused posterior.
Thus, the gains arise from using $R$ as a cross-path reference for routing and fusion.

\subsection{Anchor Utility Is Not Standalone Accuracy}
\label{sec:anchor-analysis}

Table~\ref{tab:anchor-merged} reports standalone GTPA@1 and frozen-head fusion performance when each candidate distribution is used as the shared anchor.
The central pattern is that $R$ is often weaker as a standalone predictor, yet more useful as an anchor.
$MeanF$ denotes the equal-weight average of the in-pool forward posteriors $\{F_i\}$ and is used as a replacement anchor in Table~\ref{tab:anchor-merged}.
$GenF$ is a generic pool-external single agent: one forward call with a neutral persona and the same top-$5$ output schema as the pool agents.
We use it in two roles: as a standalone baseline to test whether a generic single agent can replace collective aggregation, and as a replacement anchor to test whether a pool-external forward posterior can substitute for $R$.

\begin{table*}[t]
  \centering
  \scriptsize
  \setlength{\tabcolsep}{3pt}
  \begin{tabular}{llrrrrrr}
    \toprule
    & & \multicolumn{3}{c}{All} & \multicolumn{3}{c}{Disagree} \\
    \cmidrule(lr){3-5} \cmidrule(lr){6-8}
    \textbf{Model} & \textbf{Anchor} & Stand. & FwdJS & LogLin & Stand. & FwdJS & LogLin \\
    \midrule
    \multirow{3}{*}{Qwen3-30B}
    & $R$ & 61.50 & \textbf{73.85} & \textbf{74.25} & 40.90 & \textbf{51.58} & \textbf{52.78} \\
    & $MeanF$ & \textbf{72.69} & 72.54 & 72.69 & \textbf{48.12} & 47.67 & 48.12 \\
    & $GenF$ & 68.47 & 72.94 & 72.44 & 40.30 & 48.87 & 48.27 \\
    \midrule
    \multirow{3}{*}{Mistral-S.3.1-24B}
    & $R$ & 59.96 & \textbf{73.51} & \textbf{74.58} & 40.82 & \textbf{51.39} & \textbf{53.90} \\
    & $MeanF$ & \textbf{73.26} & 73.00 & 73.10 & \textbf{50.73} & 50.07 & 50.33 \\
    & $GenF$ & 72.24 & 72.90 & 73.10 & 48.75 & 49.80 & 50.20 \\
    \midrule
    \multirow{3}{*}{Llama-3.1-8B}
    & $R$ & 31.50 & \textbf{58.15} & \textbf{58.46} & 30.63 & \textbf{50.08} & \textbf{50.90} \\
    & $MeanF$ & \textbf{57.90} & 57.45 & 57.60 & \textbf{49.70} & 49.02 & 49.25 \\
    & $GenF$ & 46.64 & 57.60 & 56.28 & 35.96 & 49.25 & 47.52 \\
    \midrule
    \multirow{3}{*}{GLM-4-32B}
    & $R$ & 57.34 & \textbf{66.18} & \textbf{67.04} & \textbf{40.68} & \textbf{42.60} & \textbf{44.41} \\
    & $MeanF$ & \textbf{65.12} & 65.47 & 65.47 & 40.23 & 41.02 & 41.02 \\
    & $GenF$ & 61.48 & 65.88 & 64.41 & 34.80 & 41.92 & 39.55 \\
    \midrule
    \multirow{3}{*}{DeepSeek-V4-Flash}
    & $R$ & 67.76 & \textbf{78.61} & \textbf{79.07} & 36.59 & \textbf{50.81} & \textbf{52.30} \\
    & $MeanF$ & \textbf{77.80} & 77.90 & 77.80 & \textbf{48.44} & 48.74 & 48.44 \\
    & $GenF$ & 75.22 & 77.65 & 77.75 & 43.85 & 48.00 & 48.44 \\
    \bottomrule
  \end{tabular}
  \caption{Anchor utility on the headline pool (GTPA@1 \%).
  Each row uses one anchor distribution for standalone prediction and for frozen FwdJS / LogLin ($w_R{=}0.2$).
  Bold: best of $\{R,\mathrm{MeanF},\mathrm{GenF}\}$ in that column, per backbone.
  $GenF$: general single agent (pool-external, neutral persona).
  Stand.: standalone anchor argmax.}
  \label{tab:anchor-merged}
\end{table*}

On \emph{All}, $R$ is the weakest standalone predictor among $\{R,\mathrm{MeanF},\mathrm{GenF}\}$ for every backbone, trailing $MeanF$ by $7.8$--$26.4$~pp, yet using $R$ as the shared anchor yields the highest FwdJS and LogLin performance.
On \emph{Disagree}, $R$ is the weakest of the three on Mistral, Llama, and DeepSeek. GLM is the exception ($40.68\%$ for $R$ vs.\ $40.23\%$ for $MeanF$), while on Qwen $GenF$ ($40.30\%$) is slightly weaker than $R$ ($40.90\%$).
$GenF$ is often a stronger standalone predictor than $R$, yet performs worse as a replacement anchor.
Using $MeanF$ as the frozen LogLin anchor yields $\Delta{\le}0$ relative to the strongest forward-only electoral baseline on every backbone, whereas $R$ remains $+1.2$ to $+4.7$~pp above the same electoral baseline.
The electoral gain is therefore attributable to the reverse channel, not the log-linear operator alone.
Replacing $R$ with either $MeanF$ or $GenF$ lowers both fusion heads on every backbone.

The reverse construction $R(d)\propto P(e\mid d)\,P(d\mid a)$ factorizes into a likelihood term and a contextual prior. Neither one-factor variant matches the full shared $R$ on \emph{Disagree} LogLin for Qwen, Mistral, GLM, or DeepSeek. On Llama \emph{Disagree}, $R_{\mathrm{prior}}$ slightly exceeds $R$ ($51.69$ vs.\ $50.90$), but remains weaker than $MeanF$ and $GenF$ as a standalone predictor.
Appendix~\ref{sec:appendix-anchor-merged} extends Table~\ref{tab:anchor-merged} with the one-factor reverse anchors $R_{\mathrm{lik}}$ (likelihood only) and $R_{\mathrm{prior}}$ (contextual prior only).

\subsection{What the Reverse Anchor Adds Beyond the Forward Pool}
\label{sec:reverse-signal}
\label{sec:js-rank}
\label{sec:complementarity}

The two analyses below examine complementary aspects of what the reverse anchor $R$ contributes beyond the forward pool.
The first measures how often a reference predictor reproduces the forward consensus' same incorrect label, while the second evaluates whether distance to the reference provides a useful signal for ranking the five forward agents.

\paragraph{Reverse errors collide less often on the same incorrect label.}
When plurality top-1 and a reference predictor are both wrong, $\pi$ denotes the conditional probability that they assign the same incorrect label.
Table~\ref{tab:cooccur} reports three matched comparisons on the same headline pool: plurality versus $R$, versus the pool-external $GenF$, and versus each in-pool agent $F_i$, with results pooled across the five agents.
Across all five backbones, $\pi(F,R)$ is the lowest of the three ($0.196$--$0.413$), below $\pi(F,GenF)$ ($0.594$--$0.765$) and $\pi(F,F_i)$ ($0.680$--$0.829$).
The same ordering holds on \emph{Disagree} (Appendix~\ref{sec:appendix-cooccur-disagree}).
$GenF$ serves as an external extra-forward control, while $\pi(F,F_i)$ measures an in-pool echo rate because plurality is itself constructed from ${F_i}$.
Thus, $R$ shares the same mistaken label with the forward consensus less often than either a pool-external forward predictor or the agents that constitute that consensus.
Appendix~\ref{sec:appendix-cooccur-disagree} further reports the $2{\times}2$ correctness grids and Matthews $\phi$ coefficients for the corresponding error indicators.

\begin{table}[t]
  \centering
  \small
  \begin{tabular}{lccc}
    \toprule
    \textbf{Model} & $\pi(F,R)$ & $\pi(F,GenF)$ & $\pi(F,F_i)$ \\
    \midrule
    Qwen3-30B & \textbf{0.413} & 0.765 & 0.829 \\
    Mistral-S.3.1-24B & \textbf{0.404} & 0.725 & 0.775 \\
    Llama-3.1-8B & \textbf{0.196} & 0.594 & 0.680 \\
    GLM-4-32B & \textbf{0.348} & 0.658 & 0.750 \\
    DeepSeek-V4-Flash & \textbf{0.393} & 0.691 & 0.781 \\
    \bottomrule
  \end{tabular}
  \caption{Label-collision rates on \emph{All} (headline pool).
  $\pi{=}P(\mathrm{pred}_F{=}\mathrm{pred}_X\mid\text{both wrong})$ for plurality versus $X\in\{R,GenF,F_i\}$.
  $\pi(F,F_i)$ is pooled over the five pool agents.
  Bold: lowest $\pi$ per row.
  \emph{Disagree} $\pi$, $2{\times}2$ grids, and $\phi$ are in Appendix~\ref{sec:appendix-cooccur-disagree}.}
  \label{tab:cooccur}
\end{table}

\paragraph{JS divergence to $R$ ranks agents by accuracy.}
For each case, we rank the five forward agents by $\mathrm{JS}(F_i,\cdot)$ with respect to a given anchor and evaluate each rank~$k$ by the corresponding agent's GTPA@1.
In aggregate, agents with lower JS divergence tend to be more accurate than those with higher divergence.
We show Qwen as a representative case in Figure~\ref{fig:js-rank} using $R$, $MeanF$, and $GenF$ as anchors, and report all backbones in Appendix~\ref{sec:appendix-jsrank}.
The ranking is more consistently monotone under $R$ than under $MeanF$ or $GenF$.
Under $R$, accuracy decreases with rank on Qwen and Mistral, is nearly tied between rank-1 and rank-2 on DeepSeek, and peaks at rank-2 and rank-3 on Llama and GLM, respectively.
By comparison, the ranking under $MeanF$ is monotone only on Llama, while the other four backbones break the rank ordering.
Under $GenF$, only Mistral is monotone. On Qwen, Llama, and GLM, rank-2 achieves the highest accuracy, indicating that the agent closest to $GenF$ is not necessarily the most accurate.

\begin{figure}[t]
    \centering
    \includegraphics[width=\linewidth]{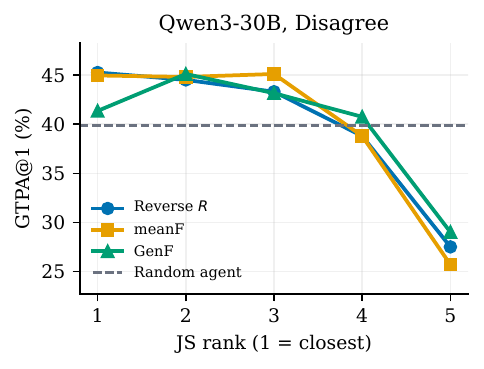}
    \caption{GTPA@1 (\%) by JS rank on \emph{Disagree} (Qwen3-30B). Reverse $R$ (blue), $MeanF$ (orange), and $GenF$ (green). Dashed line: random agent.}
    \label{fig:js-rank}
\end{figure}

\subsection{Calibrating the Reverse Anchor}
\label{sec:calib-results}
\label{sec:calibration}

Section~\ref{sec:calib-method} calibrates the reverse anchor $R$ on a labeled training split while keeping MinJS, FwdJS, and LogLin fixed.
Table~\ref{tab:calib-main} reports test GTPA@1 gains from the calibrated reverse anchor $R'$ over the uncalibrated $R$ on the same evaluation pool.
On \emph{All}, all three aggregation heads improve on all five backbones: LogLin by $0.05$--$2.83$~pp, FwdJS by $0.20$--$1.51$~pp, and MinJS by $1.06$--$4.64$~pp.
The standalone reverse top-1 accuracy also improves by $1.4$--$8.6$~pp, while the higher-capacity class-specific variant is reported separately in Appendix~\ref{sec:appendix-l1gb}.
The gains are concentrated on \emph{Disagree}: LogLin improves by $0.59$--$4.51$~pp on four backbones, with Mistral the sole exception ($-0.26$~pp).
Adding a class-specific bias $b\in\mathbb{R}^{|\mathcal{Y}|}$ further improves fusion performance, but increases reverse top-1 accuracy by up to $26$~pp on \emph{Disagree}.
This pattern is consistent with the additional capacity primarily capturing class-prior effects rather than providing a stronger routing signal.

\begin{table*}[t]
  \centering
  \small
  \setlength{\tabcolsep}{3.5pt}
  \begin{tabular}{lrrrrrrrr}
    \toprule
    & \multicolumn{4}{c}{All ($\Delta$ pp vs.\ uncalibrated)} & \multicolumn{4}{c}{Disagree ($\Delta$ pp vs.\ uncalibrated)} \\
    \cmidrule(lr){2-5} \cmidrule(lr){6-9}
    \textbf{Backbone} & $\Delta$rev & $\Delta$minJS & $\Delta$FwdJS & $\Delta$log-lin. & $\Delta$rev & $\Delta$minJS & $\Delta$FwdJS & $\Delta$log-lin. \\
    \midrule
    Qwen3-30B & $+4.97$ & $+1.86$ & $+0.80$ & $+1.61$ & $+7.22$ & $+5.41$ & $+2.41$ & $+4.51$ \\
    Mistral-S.3.1-24B & $+3.06$ & $+1.53$ & $+0.20$ & $+0.05$ & $+3.96$ & $+3.96$ & $+0.53$ & $-0.26$ \\
    Llama-3.1-8B & $+8.63$ & $+4.64$ & $+1.51$ & $+2.83$ & $+5.93$ & $+6.91$ & $+2.25$ & $+3.30$ \\
    GLM-4-32B & $+2.27$ & $+1.06$ & $+0.20$ & $+0.35$ & $+2.82$ & $+2.49$ & $+0.45$ & $+2.03$ \\
    DeepSeek-V4-Flash & $+1.42$ & $+1.22$ & $+0.25$ & $+0.10$ & $+3.56$ & $+3.41$ & $+0.74$ & $+0.59$ \\
    \bottomrule
  \end{tabular}
  \caption{Test GTPA@1 gain (pp) of calibrated $R$ over uncalibrated $R$.
  MinJS, FwdJS, and LogLin heads are frozen. The same case pools as in Tables~\ref{tab:main-valid}--\ref{tab:main-disagree} are used.}
  \label{tab:calib-main}
\end{table*}

\section{Conclusion}
\label{sec:conclusion}

LLM-based multi-agent systems aggregate heterogeneous answers, yet voting and LLM judges remain confined to the same evidence-to-conclusion path and can inherit correlated errors from the agent pool.

We introduce a shared reverse posterior $R$ by Bayesian backward reasoning as a structurally distinct reference and use $\mathrm{JS}(F_i,R)$ to guide collective decision-making.
This signal drives three training-free aggregators: hard selection (MinJS), reverse-guided reweighting (FwdJS), and lightweight forward--reverse fusion (LogLin).

Across five LLM backbones on DDXPlus, reverse-anchored aggregation improves over baseline methods, with the largest gains on \emph{Disagree}.
Notably, $R$ is often weaker as a standalone predictor, yet replacing it with either $MeanF$ or $GenF$ degrades FwdJS and LogLin, highlighting the distinction between predictive accuracy and anchor utility.
When both the forward consensus and a reference are wrong, $R$ is also less likely to share the same incorrect label with the forward consensus than $GenF$ or an in-pool agent.
Two-stage calibration further improves the frozen aggregation rules by refining the reverse anchor.
Together, these results support reverse consistency as a useful complementary signal for multi-agent collective decision-making.

\section*{Limitations}

The reverse anchor is defined over a finite label set.
It inverts an explicit likelihood $P(e\mid d)$ over a closed candidate set, using agents' shortlisted candidates, and therefore targets evidence-to-label aggregation rather than open-ended generation without a discrete candidate space.
Constructing $R$ also incurs one additional generative inversion per case on top of the $K$ forward agents.

Future work could characterize how the reliability of $R$ interacts with aggregation performance, including the point at which a sufficiently weak reverse posterior begins to hurt aggregation.
For the labeled extension, studying calibration under progressively smaller labeled sets would clarify its sample efficiency and establish how much supervision is sufficient for reliable anchor refinement.

Our evaluation focuses on a single-round, same-backbone setting.
The forward pool consists of different prompting strategies applied to one LLM backbone at a time, leaving mixed-backbone pools and multi-round debate as natural extensions.
The label-collision statistic $\pi$ captures how often two incorrect predictors assign the same wrong label, while future work could further investigate the mechanisms underlying such error overlap.
Finally, all experiments are conducted on DDXPlus, a synthetic closed-set diagnostic benchmark with 49 diseases.
Evaluating the framework across additional evidence-to-label domains would therefore be an important next step, and the proposed aggregators should not be interpreted as clinical decision-support systems.

\bibliography{references}

\appendix
\makeatletter
\setlength{\@fptop}{0pt}
\makeatother
\renewcommand{\topfraction}{0.95}
\renewcommand{\textfraction}{0.05}
\renewcommand{\floatpagefraction}{0.8}

\section{JS-based agent ranking across backbones}
\label{sec:appendix-jsrank}
Table~\ref{tab:js-rank-backbone} reports GTPA@1 after ranking the five forward agents by $\mathrm{JS}(F_i,R)$ on \emph{Disagree}.
Figure~\ref{fig:js-rank} in the main text shows the Qwen \emph{Disagree} slice and includes $MeanF$ and $GenF$ as alternative anchors.
Figure~\ref{fig:js-rank-app-all} summarizes the remaining backbone--subset curves for both \emph{All} and \emph{Disagree}.

Under $R$, the \emph{Disagree} rank curve is strictly monotone on Qwen and Mistral. Llama and GLM attain their highest accuracy after rank-1, while rank-1 and rank-2 on DeepSeek differ by only one case. The ranking is less consistently monotone under $MeanF$ and $GenF$.

\begin{table}[htbp]
  \centering
  \small
  \setlength{\tabcolsep}{3pt}
  \begin{tabularx}{\columnwidth}{>{\raggedright\arraybackslash}X c c c}
    \toprule
    \textbf{Model} & Rank-1 & Rank-5 & Rand. \\
    \midrule
    Qwen3-30B & 45.26 & 27.52 & 39.88 \\
    Mistral-S.3.1-24B & 46.24 & 29.19 & 40.42 \\
    Llama-3.1-8B & 37.76 & 26.73 & 35.36 \\
    GLM-4-32B & 38.19 & 24.63 & 34.21 \\
    DeepSeek-V4-Flash & 48.15 & 25.48 & 41.19 \\
    \bottomrule
  \end{tabularx}
  \caption{GTPA@1 (\%) by reverse JS rank on \emph{Disagree}.
  Rank-$k$: the forward agent with the $k$-th lowest $\mathrm{JS}(F_i,R)$ on each case.
  Rand.: mean GTPA@1 across the five ranks.}
  \label{tab:js-rank-backbone}
\end{table}

\begin{figure*}[t]
  \centering
  \small
  \setlength{\tabcolsep}{2pt}
  \begin{tabular}{@{}ccc@{}}
    \multicolumn{3}{@{}l}{\emph{All}} \\
    \includegraphics[width=0.25\linewidth]{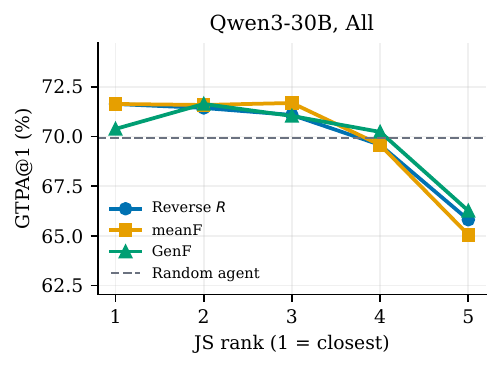} &
    \includegraphics[width=0.25\linewidth]{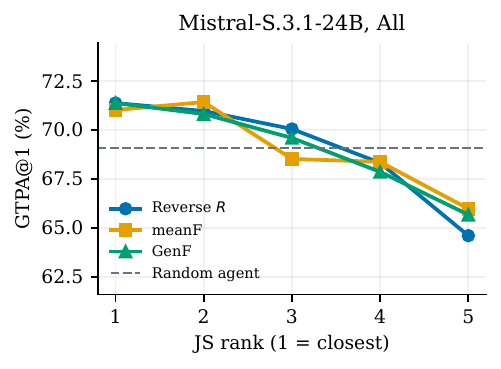} &
    \includegraphics[width=0.25\linewidth]{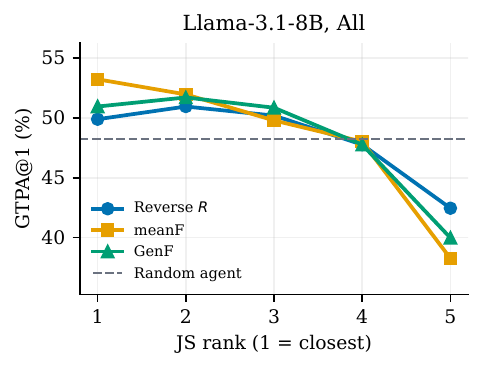} \\
    \includegraphics[width=0.25\linewidth]{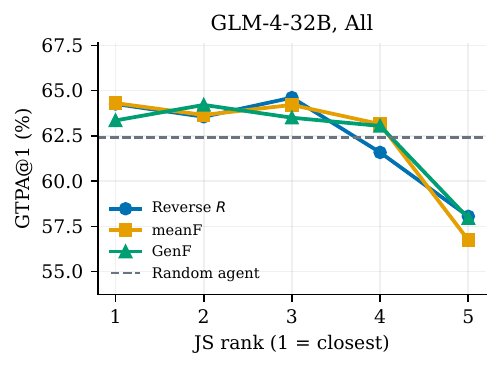} &
    \includegraphics[width=0.25\linewidth]{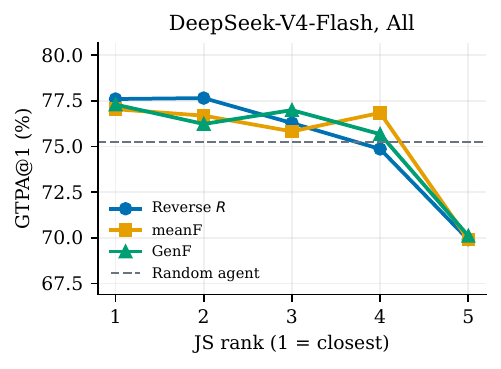} &
    \\[2pt]
    \multicolumn{3}{@{}l}{\emph{Disagree} (Qwen in Figure~\ref{fig:js-rank})} \\
    \includegraphics[width=0.25\linewidth]{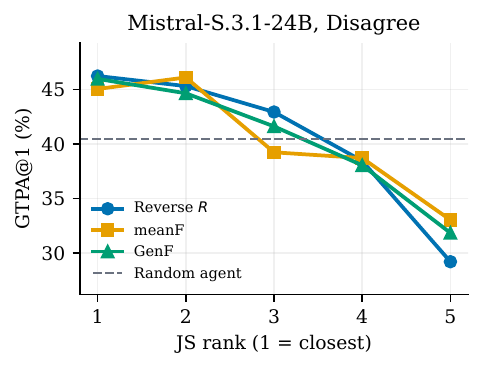} &
    \includegraphics[width=0.25\linewidth]{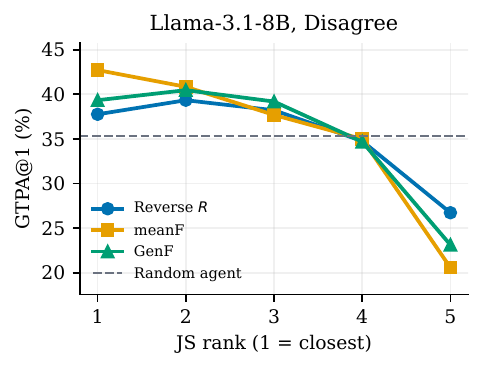} &
    \includegraphics[width=0.25\linewidth]{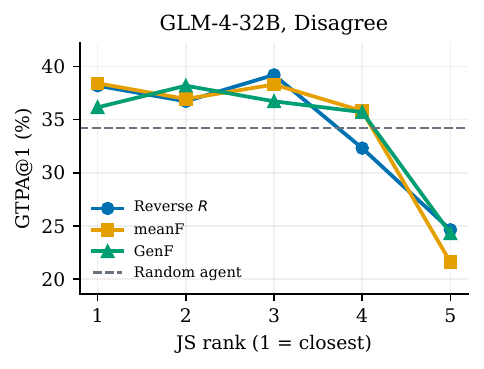} \\
    \multicolumn{3}{c}{\includegraphics[width=0.25\linewidth]{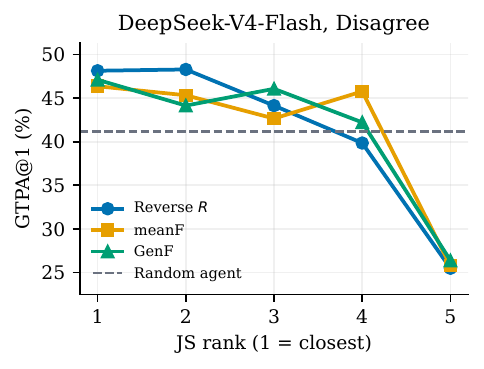}}
  \end{tabular}
  \caption{GTPA@1 (\%) by JS rank on the evaluation pool.
  Reverse $R$, $MeanF$, and $GenF$ are used as anchors.
  Dashed line: random agent.}
  \label{fig:js-rank-app-all}
  \label{fig:js-rank-app-disagree}
\end{figure*}

\section{Labeled calibration: fitting and capacity upper bound}
\label{sec:appendix-l1gb}

Section~\ref{sec:calib-method} defines the two-stage calibration of $R$.
This appendix details the fitting protocol for the calibrated reverse anchor and the higher-capacity class-specific bias variant.

\subsection{Fitting protocol}
The reverse inversion uses ordinal evidence-likelihood and context-activation bins.
Each axis in~\eqref{eq:l5curve} has seven ordinal ranks $k\in{0,\ldots,6}$, with axis-specific clamps $\ell,h$.
Stage~1 fits four map parameters $(a_s,b_s,a_a,b_a)$ together with a temperature $T$ for rescaling the replayed reverse posterior.
The objective is to minimize the label NLL of the reverse posterior $R$ obtained by replaying the Bayesian network on the training cases.
After fitting, the two maps and the temperature $T$ are frozen.
Stage~2 fits the scalar $\gamma$ in~\eqref{eq:l1g} by $k$-fold minimization of the same NLL on the training fit split.
Evaluation then applies the frozen MinJS, FwdJS, and LogLin heads on the same evaluation pool used for the label-free results.

\subsection{Capacity upper bound (\textsc{L5curve\_L1gb})}
Table~\ref{tab:calib-l1gb}  reports the same frozen-head evaluation as Section~\ref{sec:calib-results}, but gives each label its own bias term $b_d$ in addition to the scalar correction $\gamma$:
\begin{equation}
R'(d)\propto R(d)\,m(d)^{-\gamma}\,e^{b_d}.
\end{equation}
This gives the calibration more flexibility and further improves both the standalone accuracy of $R$ and the downstream fusion results.
However, the large gain in reverse top-1 accuracy suggests that much of this improvement may come from adjusting label-specific prior preferences rather than from making $R$ a better routing signal.
We therefore use this higher-capacity version only as an upper-bound reference, rather than as the primary calibrated method.

\begin{table}[htbp]
  \centering
  \scriptsize
  \setlength{\tabcolsep}{2.4pt}
  \resizebox{\columnwidth}{!}{%
  \begin{tabular}{lrrrrrrrr}
    \toprule
    & \multicolumn{4}{c}{All ($\Delta$ pp vs.\ uncalibrated)} & \multicolumn{4}{c}{Disagree ($\Delta$ pp vs.\ uncalibrated)} \\
    \cmidrule(lr){2-5} \cmidrule(lr){6-9}
    \textbf{Backbone} & $\Delta$rev & $\Delta$minJS & $\Delta$FwdJS & $\Delta$log-lin. & $\Delta$rev & $\Delta$minJS & $\Delta$FwdJS & $\Delta$log-lin. \\
    \midrule
    Qwen3-30B & $+12.05$ & $+3.71$ & $+2.06$ & $+5.77$ & $+19.10$ & $+10.98$ & $+6.17$ & $+13.23$ \\
    Mistral-S.3.1-24B & $+7.90$ & $+2.29$ & $+0.97$ & $+2.75$ & $+11.76$ & $+5.55$ & $+2.51$ & $+5.81$ \\
    Llama-3.1-8B & $+16.15$ & $+7.32$ & $+3.33$ & $+6.81$ & $+14.49$ & $+10.96$ & $+4.95$ & $+8.86$ \\
    GLM-4-32B & $+10.70$ & $+3.33$ & $+1.41$ & $+4.24$ & $+12.66$ & $+7.57$ & $+3.16$ & $+8.70$ \\
    DeepSeek-V4-Flash & $+12.52$ & $+4.36$ & $+1.32$ & $+3.75$ & $+26.52$ & $+12.59$ & $+3.85$ & $+11.11$ \\
    \bottomrule
  \end{tabular}%
  }
  \caption{Test GTPA@1 gain (pp) from class-specific calibration of $R$ over uncalibrated $R$.
  Same evaluation pool as Table~\ref{tab:calib-main}.}
  \label{tab:calib-l1gb}
\end{table}

\section{Likelihood-only and prior-only reverse anchors}
\label{sec:appendix-anchor-merged}

Table~\ref{tab:anchor-merged} in the main text compares the full reverse posterior $R$ with the forward-pool anchors $MeanF$ and $GenF$.
Table~\ref{tab:appendix-anchor-full} uses the same evaluation pools and frozen FwdJS/LogLin heads, but replaces $R$ with one-factor reverse constructions: $R_{\mathrm{lik}}$ uses only the likelihood term $P(e\mid d)$, whereas $R_{\mathrm{prior}}$ uses only the contextual prior $P(d\mid a)$.
On \emph{Disagree} with LogLin, neither one-factor construction matches the full $R$ on Qwen, Mistral, GLM, or DeepSeek. Llama is the exception: $R_{\mathrm{prior}}$ reaches $51.69\%$, slightly above $R$ at $50.90\%$, but remains weaker than $MeanF$ and $GenF$ as a standalone predictor.

\begin{table}[htbp]
  \centering
  \scriptsize
  \setlength{\tabcolsep}{2.2pt}
  \resizebox{\columnwidth}{!}{%
  \begin{tabular}{llrrrrrr}
    \toprule
    & & \multicolumn{3}{c}{All} & \multicolumn{3}{c}{Disagree} \\
    \cmidrule(lr){3-5} \cmidrule(lr){6-8}
    \textbf{Model} & \textbf{Anchor} & Stand. & FwdJS & LogLin & Stand. & FwdJS & LogLin \\
    \midrule
    \multirow{2}{*}{Qwen3-30B} & $R_{\mathrm{lik}}$ & 51.53 & 72.90 & 72.90 & 30.23 & 48.87 & 48.87 \\
    & $R_{\mathrm{prior}}$ & 43.69 & 73.20 & 73.40 & 31.58 & 49.77 & 50.38 \\
    \midrule
    \multirow{2}{*}{Mistral-S.3.1-24B} & $R_{\mathrm{lik}}$ & 44.74 & 73.06 & 74.23 & 34.44 & 50.20 & 53.11 \\
    & $R_{\mathrm{prior}}$ & 43.21 & 73.21 & 73.83 & 26.23 & 50.60 & 52.19 \\
    \midrule
    \multirow{2}{*}{Llama-3.1-8B} & $R_{\mathrm{lik}}$ & 19.67 & 57.89 & 56.88 & 16.98 & 49.74 & 48.53 \\
    & $R_{\mathrm{prior}}$ & 30.59 & 58.29 & 59.20 & 29.53 & 50.34 & 51.69 \\
    \midrule
    \multirow{2}{*}{GLM-4-32B} & $R_{\mathrm{lik}}$ & 47.32 & 64.86 & 65.12 & 33.75 & 39.64 & 40.32 \\
    & $R_{\mathrm{prior}}$ & 39.89 & 65.57 & 66.03 & 25.37 & 41.22 & 42.13 \\
    \midrule
    \multirow{2}{*}{DeepSeek-V4-Flash} & $R_{\mathrm{lik}}$ & 59.49 & 78.02 & 77.87 & 32.89 & 49.19 & 48.74 \\
    & $R_{\mathrm{prior}}$ & 48.93 & 77.92 & 78.58 & 23.41 & 48.89 & 50.81 \\
    \bottomrule
  \end{tabular}%
  }
  \caption{One-factor reverse anchors (GTPA@1 (\%)) on the same evaluation pools as Table~\ref{tab:anchor-merged}.
  $R_{\mathrm{lik}}$: likelihood-only, using $P(e\mid d)$; $R_{\mathrm{prior}}$: prior-only, using $P(d\mid a)$.}
  \label{tab:appendix-anchor-full}
\end{table}

\section{Log-linear $w_R$ sweep}
\label{sec:appendix-wr-sweep}

Table~\ref{tab:wr-sweep} reports LogLin GTPA@1 for $w_R\in{0.2,0.3,0.4,0.5}$ on the same evaluation pools as Tables~\ref{tab:main-valid}, \ref{tab:main-disagree}.
The $w_R{=}0.2$ column matches the LogLin results reported in the main tables.
Because FwdJS has already incorporated $R$ through JS-based reweighting, we use $w_R{=}0.2$ as a light direct contribution from the reverse anchor rather than selecting the weight separately for each backbone on the test set.
On \emph{Disagree} subset, larger $w_R$ can yield higher accuracy on some backbones. For example, GLM improves from $44.41\%$ at $w_R{=}0.2$ to $47.57\%$ at $w_R{=}0.5$.

\begin{table}[htbp]
  \centering
  \small
  \setlength{\tabcolsep}{3pt}
  \resizebox{\columnwidth}{!}{%
  \begin{tabular}{lrrrrrrrr}
    \toprule
    & \multicolumn{4}{c}{All} & \multicolumn{4}{c}{Disagree} \\
    \cmidrule(lr){2-5} \cmidrule(lr){6-9}
    \textbf{Model} & $0.2$ & $0.3$ & $0.4$ & $0.5$ & $0.2$ & $0.3$ & $0.4$ & $0.5$ \\
    \midrule
    Qwen3-30B & \textbf{74.25} & \textbf{74.25} & \textbf{74.25} & 73.85 & 52.78 & 53.08 & \textbf{53.38} & 53.08 \\
    Mistral-Small-3.1-24B & 74.58 & \textbf{74.73} & 73.71 & 73.10 & 53.90 & \textbf{54.29} & 52.97 & 51.78 \\
    Llama-3.1-8B & \textbf{58.46} & 57.40 & 55.83 & 53.36 & \textbf{50.90} & 49.92 & 49.40 & 47.97 \\
    GLM-4-32B & \textbf{67.04} & 66.99 & 66.89 & 66.89 & 44.41 & 45.31 & 45.88 & \textbf{47.57} \\
    DeepSeek-V4-Flash & \textbf{79.07} & 79.02 & 78.97 & 78.71 & \textbf{52.30} & 52.00 & 52.15 & 52.00 \\
    \bottomrule
  \end{tabular}%
  }
  \caption{Log-linear GTPA@1 (\%) as a function of reverse weight $w_R$ on the evaluation pools. Bold denotes the best $w_R$ within each block. The main tables use the fixed default $w_R{=}0.2$.}
  \label{tab:wr-sweep}
\end{table}

\section{Label collision, co-occurrence, and error-indicator $\phi$}
\label{sec:appendix-cooccur-disagree}

Table~\ref{tab:appendix-cooccur-disagree} reports the matched-unit $\pi$ statistic from Table~\ref{tab:cooccur} on \emph{Disagree} subset.
Across all five backbones, $\pi(F,R)$ remains the lowest of the three comparisons, consistent with the \emph{All} results in the main text.
Tables~\ref{tab:appendix-grid} and~\ref{tab:appendix-phi} provide complementary views that are not used for the main same-incorrect-label claim: a $2{\times}2$ correctness grid and the Matthews $\phi$ coefficient computed from binary error indicators.
The grid uses the same plurality GTPA@1 indicator as Tables~\ref{tab:main-valid}--\ref{tab:main-disagree}, so Both$\checkmark$+$F\checkmark R\times$ matches the plurality column of those tables.
Here, $\phi$ measures the association between whether plurality is wrong and whether the named reference is wrong. Unlike $\pi$, it does not measure whether the two predictors assign the same incorrect label.

\begin{table}[htbp]
  \centering
  \small
  \begin{tabular}{lccc}
    \toprule
    \textbf{Model} & $\pi(F,R)$ & $\pi(F,GenF)$ & $\pi(F,F_i)$ \\
    \midrule
    Qwen3-30B & \textbf{0.333} & 0.639 & 0.717 \\
    Mistral-S.3.1-24B & \textbf{0.332} & 0.621 & 0.677 \\
    Llama-3.1-8B & \textbf{0.162} & 0.497 & 0.600 \\
    GLM-4-32B & \textbf{0.296} & 0.588 & 0.663 \\
    DeepSeek-V4-Flash & \textbf{0.352} & 0.609 & 0.716 \\
    \bottomrule
  \end{tabular}
  \caption{Label-collision rates on \emph{Disagree}.
  Definitions match Table~\ref{tab:cooccur}.
  Bold denotes the lowest $\pi$ in each row.}
  \label{tab:appendix-cooccur-disagree}
\end{table}

\begin{table}[htbp]
  \centering
  \scriptsize
  \setlength{\tabcolsep}{2.4pt}
  \resizebox{\columnwidth}{!}{%
  \begin{tabular}{lcccccccc}
    \toprule
    & \multicolumn{4}{c}{All} & \multicolumn{4}{c}{Disagree} \\
    \cmidrule(lr){2-5} \cmidrule(lr){6-9}
    \textbf{Model} &
    Both$\checkmark$ & F$\checkmark$R$\times$ & F$\times$R$\checkmark$ & Both$\times$ &
    Both$\checkmark$ & F$\checkmark$R$\times$ & F$\times$R$\checkmark$ & Both$\times$ \\
    \midrule
    Qwen3-30B & 50.8 & 20.8 & 10.5 & 17.9 & 22.6 & 22.4 & 20.3 & 34.7 \\
    Mistral-S.3.1-24B & 50.8 & 20.6 & 9.6 & 19.1 & 25.4 & 20.9 & 17.7 & 36.1 \\
    Llama-3.1-8B & 21.6 & 32.6 & 11.1 & 34.8 & 18.0 & 26.1 & 13.8 & 42.0 \\
    GLM-4-32B & 45.3 & 20.0 & 13.3 & 21.4 & 21.4 & 19.4 & 21.4 & 37.9 \\
    DeepSeek-V4-Flash & 62.0 & 14.9 & 6.8 & 16.4 & 23.4 & 22.5 & 15.6 & 38.5 \\
    \bottomrule
  \end{tabular}%
  }
  \caption{Forward/reverse correctness co-occurrence (\% of the slice).
  Columns denote Both$\checkmark$, plurality correct and $R$ wrong, plurality wrong and $R$ correct (recovery), and Both$\times$.
  Plurality correctness is the same GTPA@1 indicator as in Tables~\ref{tab:main-valid}--\ref{tab:main-disagree}.}
  \label{tab:appendix-grid}
\end{table}

\begin{table}[htbp]
  \centering
  \small
  \setlength{\tabcolsep}{3pt}
  \resizebox{\columnwidth}{!}{%
  \begin{tabular}{lcccccc}
    \toprule
    & \multicolumn{3}{c}{All} & \multicolumn{3}{c}{Disagree} \\
    \cmidrule(lr){2-4} \cmidrule(lr){5-7}
    \textbf{Model} & $\phi(F,R)$ & $\phi(F,GenF)$ & $\phi(F,F_i)$ & $\phi(F,R)$ & $\phi(F,GenF)$ & $\phi(F,F_i)$ \\
    \midrule
    Qwen3-30B & \textbf{0.31} & 0.73 & 0.81 & \textbf{0.13} & 0.41 & 0.53 \\
    Mistral-S.3.1-24B & \textbf{0.35} & 0.79 & 0.81 & \textbf{0.24} & 0.61 & 0.60 \\
    Llama-3.1-8B & \textbf{0.16} & 0.61 & 0.71 & \textbf{0.17} & 0.45 & 0.56 \\
    GLM-4-32B & \textbf{0.30} & 0.74 & 0.82 & \textbf{0.16} & 0.53 & 0.60 \\
    DeepSeek-V4-Flash & \textbf{0.47} & 0.71 & 0.78 & \textbf{0.23} & 0.46 & 0.52 \\
    \bottomrule
  \end{tabular}%
  }
  \caption{Matthews $\phi$ on binary error indicators.
  Each $\phi$ measures the association between whether plurality is wrong and whether the named reference is wrong, rather than whether they assign the same incorrect label.
  Unlike Table~\ref{tab:appendix-grid}, the error indicators here use string-level top-1 equality with the gold label, not plurality GTPA@1.
  $\phi(F,F_i)$ is pooled across the five in-pool agents.
  Bold denotes the lowest $\phi$ within each \emph{All} or \emph{Disagree} block.}
  \label{tab:appendix-phi}
\end{table}

\end{document}